\documentclass[conference]{IEEEtran}
\IEEEoverridecommandlockouts
\usepackage{cite}
\usepackage{amsmath,amssymb,amsfonts}
\usepackage{algorithmic}
\usepackage{graphicx}
\usepackage{textcomp}
\usepackage{xcolor}
\usepackage{booktabs}
\usepackage{makecell}
\usepackage{url}
\def\BibTeX{{\rm B\kern-.05em{\sc i\kern-.025em b}\kern-.08em
    T\kern-.1667em\lower.7ex\hbox{E}\kern-.125emX}}

\usepackage{caption}

\begin{document}

\title{Improving Buoy Detection with Deep Transfer Learning for Mussel Farm Automation\\
{}
\thanks{979-8-3503-7051-5/23/\$31.00 ©2023 IEEE}
}

\author{\IEEEauthorblockN{ Carl McMillan\IEEEauthorrefmark{1}, Junhong Zhao\IEEEauthorrefmark{1}, Bing Xue\IEEEauthorrefmark{1}, Ross Vennell\IEEEauthorrefmark{2}, Mengjie Zhang\IEEEauthorrefmark{1}}
\IEEEauthorblockA{\IEEEauthorrefmark{1}\textit{Centre for Data Science and Artificial Intelligence \& School of Engineering and Computer Science } \\
\textit{Victoria University of Wellington, New Zealand} \\
\IEEEauthorrefmark{2}\textit{Coastal and Freshwater Group, Cawthron Institute, Nelson, New Zealand} \\
Email: carl.mcmillan@ecs.vuw.ac.nz; j.zhao@vuw.ac.nz; bing.xue@ecs.vuw.ac.nz; \\ ross.vennell@cawthron.org.nz; mengjie.zhang@ecs.vuw.ac.nz
}}

\maketitle

\begin{abstract}
The aquaculture sector in New Zealand is experiencing rapid expansion, with a particular emphasis on mussel exports. As the demands of mussel farming operations continue to evolve, the integration of artificial intelligence and computer vision techniques, such as intelligent object detection, is emerging as an effective approach to enhance operational efficiency. This study delves into advancing buoy detection by leveraging deep learning methodologies for intelligent mussel farm monitoring and management. The primary objective centers on improving accuracy and robustness in detecting buoys across a spectrum of real-world scenarios. A diverse dataset sourced from mussel farms is captured and labeled for training, encompassing imagery taken from cameras mounted on both floating platforms and traversing vessels, capturing various lighting and weather conditions. To establish an effective deep learning model for buoy detection with a limited number of labeled data, we employ transfer learning techniques. This involves adapting a pre-trained object detection model to create a specialized deep learning buoy detection model. We explore different pre-trained models, including YOLO and its variants, alongside data diversity to investigate their effects on model performance. Our investigation demonstrates a significant enhancement in buoy detection performance through deep learning, accompanied by improved generalization across diverse weather conditions, highlighting the practical effectiveness of our approach.
\end{abstract}


\section{Introduction}

New Zealand (NZ) green-lipped mussels (Perna Canaliculus) make up the largest part of NZ's aquaculture exports and the NZ government is investing in growing the aquaculture industry \cite{ministry_for_primary_industries_nz_government_new_2019}. However, traditional ways of monitoring mussel farm lines can be costly due to the manual work involved, particularly for buoy inspection, where factors like buoy loss or submergence introduce complexities. AI-based buoy detection offers a promising solution to help automate the monitoring process. Cameras in the mussel farm could discern buoy positions with intelligent real-time analysis of the video feed.

Previous studies on buoy detection using keypoint techniques have been performed on mussel farms in the Marlborough Sounds, NZ \cite{Bi2022,zeng_new_2023}. They focus on identifying distinctive key points on buoys, a step towards frame alignment. Although keypoint-based solutions exhibit effectiveness with high-quality images, they struggle with challenging conditions, such as the presence of strong capillary waves in the water or low contrast, demonstrating the constrained robustness for handling in-the-wild scenarios \cite{Bi2022,zeng_new_2023}. It prompts the exploration of alternative strategies like deep learning techniques, such as leveraging convolutional neural networks (CNNs), to advance the applicability of this method to real use.

In this study, we develop a deep learning model for real time detection of mussel farm buoys in a variety of conditions to predict bounding boxes around the buoys. Initially, a dataset was curated, featuring precise annotations of buoys in diverse sizes within the image. Given the restricted volume of this annotated data, we resort to employing transfer learning to leverage pre-trained models for effective weight initialization of our model and to address the demand for data mitigation. 

We use a recent YOLO (You Only Look Once) implementation, YOLOv7 \cite{wang_yolov7_2022}, a deep learning-based single-stage object detector as our pre-trained model. Through transfer learning, we fine-tune the model's weights on our buoy image dataset, tailoring it to the specific buoy detection task.  YOLO can learn the global semantic context of image content through multiple convolutional operations across the entire image, effectively predicting bounding boxes of the target object in real time \cite{redmon_you_2016}. This semantic feature inference is essential for buoy detection, potentially enabling the model to discern buoy line patterns and making it more likely to predict buoys in those areas, thus mitigating detection errors \cite{redmon_you_2016}. In addition, YOLO excels in generalizing detection across various image styles and variations\cite{redmon_you_2016}, which is particularly applicable to the diverse weather conditions encountered in the buoy detection task. In the mussel farming usage scenario, a critical element that needs to be considered is the mobility and real-time applicability of the developed model during inference. YOLO has a relatively small architecture and can process frames quickly due to being a single-stage object detector \cite{redmon_you_2016}, demonstrating it to be advantageous in addressing these issues.

Benefiting from the high performance of YOLO, our buoy detection model exhibits a substantial enhancement in detection accuracy and resilience in addressing a wide array of in-the-wild conditions prevalent in mussel farms. Furthermore, the model operates in real time, is capable of execution on a standard GPU device and could be integrated into systems requiring real time buoy detection. This marks a further step towards the realization of intelligent mussel farming.

\section{Related Work}

Several studies have investigated AI-based buoy detection in NZ mussel farms \cite{Bi2022,zeng_new_2023}. These studies segment the images using a U-Net to remove land above the waterline, then use traditional computer vision techniques such as local binary patterns or laplacian of Gaussians combined with keypoint detection to find and classify keypoint descriptors on the buoys. These methods do not find bounding boxes around the buoys but are useful for finding keypoints which could be used later for matching between different mussel farm images.

Zhao et al. \cite{zhao_yolov7-sea_2023} investigate object detection in the sea for search and rescue and ship navigation using YOLOv7 on a large sea image dataset called SeaDronesee. They have difficulty with small object detection on this dataset due to large areas with small objects. They make improvements to the YOLOv7 model for their use case by adding a prediction head to detect small objects as well as an attention module to focus on regions in the large scene. They also improve image augmentations. Their method achieved a 7\% improvement in mAP (mean average precision) over standard YOLOv7.

Recent studies have used YOLOv3 \cite{redmon_yolov3_2018} to detect objects in adverse weather showing good results \cite{liu_image-adaptive_2022,wang_denet_2023}. These studies use preprocessing models before feeding into the YOLO model. Liu et al. \cite{liu_image-adaptive_2022} use a CNN model to determine parameters for an image pre-processing pipeline before feeding into YOLOv3 for detection. Wang et al. \cite{wang_denet_2023} apply Laplacian pyramid decomposition to extract low and high frequency representations of the image, recomposing these into a new image that achieves better detection results.

While these studies \cite{liu_image-adaptive_2022,wang_denet_2023} add pre-processing to improve the quality of images containing adverse weather before performing detection using YOLOv3, Haimer et al. \cite{haimer_yolo_2023} compare various YOLO versions without additional pre-processing. They compare YOLOv3, YOLOv4, YOLOv5, YOLOv6, and YOLOv7 in traffic images. Their results showed that YOLOv7 outperformed all other tested YOLO versions for both accuracy and speed on images with adverse weather conditions.

\section{Method}

\subsection{Algorithm Overview}

Our study resorts to deep learning for buoy detection due to the impressive performance of deep learning models for general object detection. However, training deep learning models for object detection from scratch typically takes significant time and computational expense, demanding a large amount of annotated images. Given the constraints of a limited number of annotated buoy images, we opted for transfer learning to fine-tune a pre-trained model using our limited buoy dataset.  

YOLO was selected over other deep learning algorithms due to its ability to learn global context for reducing false positives \cite{redmon_you_2016} and its ability to generalise \cite{redmon_you_2016} which could be useful for handling the range of weather conditions encountered in mussel farms. It is also known for its fast detection speeds \cite{redmon_you_2016}. Specifically, we explored the state-of-the-art YOLOv7 model \cite{wang_yolov7_2022}, which has been used in other relevant studies \cite{zhao_yolov7-sea_2023,haimer_yolo_2023}.

\begin{figure}[htbp]
	\centerline{\includegraphics[width=8.8cm]{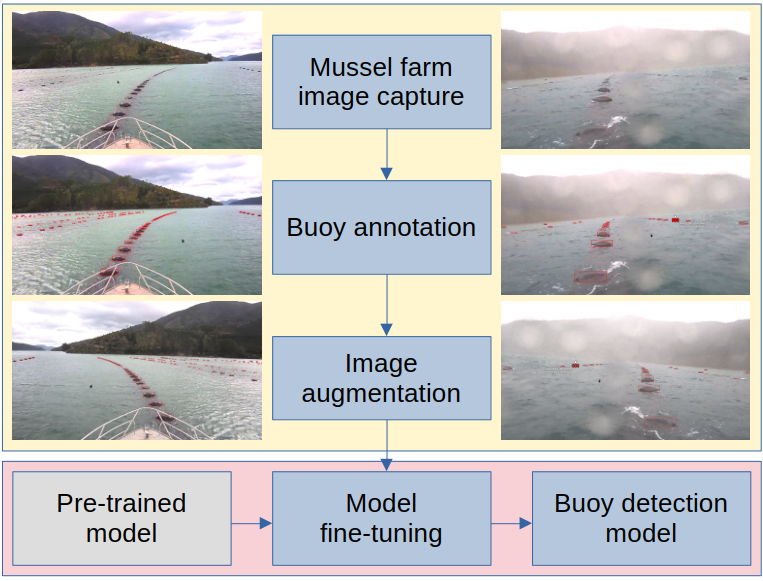}}
	\caption{Raw images are collected from the cameras, then annotated with buoy bounding boxes. Image augmentations are applied to the image dataset before fine-tuning the pre-trained model, resulting in our buoy detection model.}
	\label{fig:learning_pipeline}
\end{figure}

This investigation evaluates the standard YOLOv7 model and its counterpart, the YOLOv7-tiny model, featuring a smaller network architecture designed for faster inference on portable devices \cite{wang_yolov7_2022}. The pre-trained models used for weights initialization were trained on the Microsoft COCO image dataset with diverse object categories, including cars, birds, traffic lights, etc. Transfer learning was performed on the pre-trained models using our unique mussel farm image datasets, which were annotated with bounding boxes around the buoys in diverse sizes, reflectance, and colour variations. To improve the effectiveness of the final model, we performed data augmentation on our labelled dataset. All image augmentation operations we adopted are listed in Table~\ref{table:training_params}.  The hyperparameters for model training are included in Table \ref{table:training_params}. The chosen values were suggested from the YOLOv7 source code repository \cite{yolov7_hyperparameters}. Future work may be undertaken to further optimise these parameters. The outcome resulted in a fine-tuned model tailored specifically for buoy detection. The proposed pipeline is illustrated in Fig. \ref{fig:learning_pipeline}.

\begin{table}[htbp]
	\caption{Training parameters}
	\centering
	\begin{tabular}[t]{ l c l c }
		\toprule
        \multicolumn{2}{c}{Hyperparameters} & \multicolumn{2}{c}{Image Augmentation} \\
        \cmidrule(lr){1-2}\cmidrule(lr){3-4}
		Parameter & Value & Parameter & Value \\ [0.25ex]
		\cmidrule(lr){1-2}\cmidrule(lr){3-4}
		Training epochs                 & 50     & Focal loss gamma            & 0.0    \\
        Batch size                      & 8      & HSV-Hue                     & 0.015  \\
        Number of workers               & 8      & HSV-Saturation              & 0.7    \\
        Optimizer                       & SGD    & HSV-Value                   & 0.4    \\
		Initial learning rate           & 0.01   & Rotation degrees            & 0.0    \\
        Final learning rate             & 0.01   & Translation                 & 0.1    \\
        SGD Momentum                    & 0.937  & Scale                       & 0.5    \\
        Optimizer weight decay          & 0.0005 & Shear                       & 0.0    \\
        Warmup epochs                   & 3.0    & Perspective                 & 0.0    \\
        Warmup momentum                 & 0.8    & Flip up-down prob.          & 0.0    \\
        Warmup bias learning rate       & 0.1    & Flip left-right prob.       & 0.5    \\
        Box loss gain                   & 0.05   & Mosaic prob.                & 1.0    \\
        Class loss gain                 & 0.5    & Mixup prob.                 & 0.05   \\
        Object loss gain                & 1.0    & \\
        Anchor target threshold         & 4.0    & \\
		\bottomrule
	\end{tabular}
	\label{table:training_params}
\end{table}

\subsection{Mussel Farm Dataset Establishment}

The image datasets for this study were captured from three camera sources deployed in the mussel farm.

\subsubsection{Boat camera}
These images are from cameras mounted on a boat. Video was taken while the boats were traversing the mussel farm. A random selection of frames was taken from videos. The raw frame resolution was 1920x1080 pixels. The images are typically in fine weather during daylight, showing high-quality image details. See Row 1 in Fig. \ref{fig:image_samples} for samples.

\subsubsection{Low-resolution buoy mounted camera}
These images are from a camera mounted on a buoy in the mussel farm. Videos were taken in all weather conditions and all light conditions (see Fig.~\ref{fig:image_samples}, 2nd row). A random selection of frames was taken from videos, and images that were too dark were removed. The raw frame resolution was 1920x1080 pixels. 

\subsubsection{High-resolution buoy mounted camera}
\begin{figure}[htbp]
	\centerline{\includegraphics[width=8.8cm]{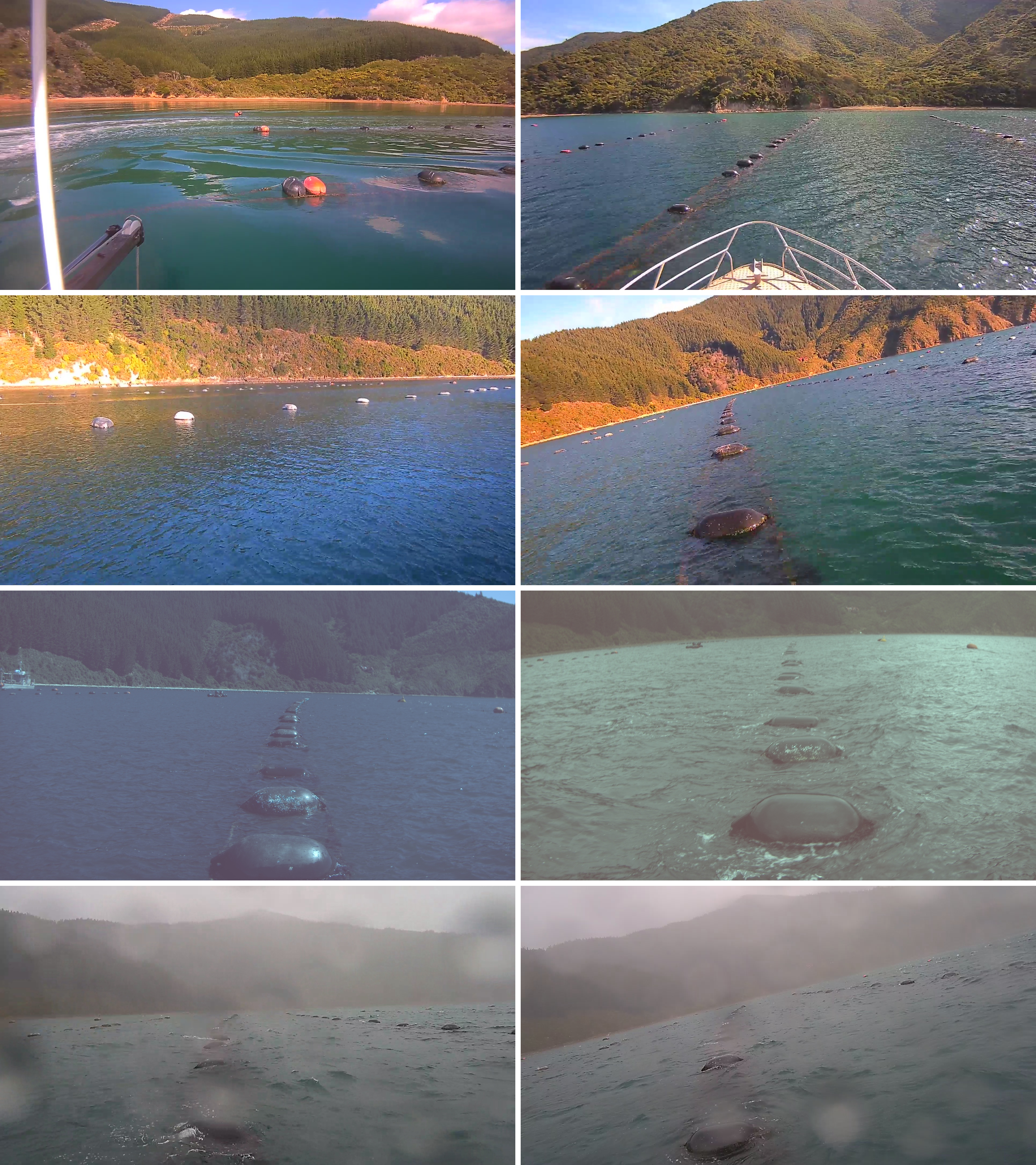}}
	\caption{Image samples. Row 1: Boat camera. Row 2: Low-res buoy-mounted camera. Row 3: High-res buoy-mounted camera. Row 4: Adverse weather low-res buoy-mounted camera.}
	\label{fig:image_samples}
\end{figure}

These images are from a higher-resolution camera mounted on a buoy in the mussel farm. Videos were taken in all weather conditions and all light conditions (see Fig.~\ref{fig:image_samples}, 3rd row). A random selection of frames was taken from videos, filtering out images that were too dark. The raw frame resolution was 4032x3040 pixels.

The datasets were carefully ablated to explore their impacts on the final performance. Furthermore, an additional adverse weather test set was established from the low-resolution buoy-mounted camera source through a random selection of frames in a video where the weather was very windy, with limited visibility and rain or sea spray in the air hitting the camera (see Fig.~\ref{fig:image_samples}, 4th row). Using it, we aim to test the robustness of our model on images with challenging in-the-wild conditions.

All images were annotated with bounding boxes around the buoys that could be distinguished. All datasets except for adverse weather were split into training, validation, and testing at a 70/10/20 ratio. The adverse weather dataset was only used for testing. The resulting dataset sizes are in Table \ref{table:datasets}.

\begin{table}[htbp]
	\caption{Datasets}
	\centering
	\begin{tabular}[t]{ l r r r }
		\toprule
		Dataset & \thead{Train\\size} & \thead{Validation\\size} & \thead{Test\\size} \\ [0.25ex]
		\midrule
		Boat camera                    & 489 & 70  & 141 \\
		Buoy mounted camera (low res)  & 112 & 16  & 32  \\
		Buoy mounted camera (high res) & 126 & 18  & 37  \\
		All combined                   & 728 & 104 & 209 \\
        Adverse weather                & 0   & 0   & 50  \\
		\bottomrule
	\end{tabular}
	\label{table:datasets}
\end{table}

\subsection{Model fine-tune}

We fine-tuned YOLOv7 and its variants~\cite{wang_yolov7_2022} with mussel farm image datasets. The YOLOv7 \cite{wang_yolov7_2022} implementation was cloned from the official implementation \cite{yolov7_github}. Weights of YOLOv7~\cite{yolov7_pretrained_full} and its counterpart YOLOv7-tiny~\cite{yolov7_pretrained_tiny} pre-trained on the Microsoft COCO dataset \cite{lin_microsoft_2014} were used as starting points for training with the buoy image data. We retain all convolutional layers as trainable layers and substitute the number of output classes per bounding box anchor to one class of buoy. All models were trained using the same hyperparameters~\cite{yolov7_hyperparameters} and image augmentations shown in Table \ref{table:training_params}. Four variations of the models from Table \ref{table:models} were trained over 50 epochs each, with batch sizes of 8.

\begin{table}[htbp]
\caption{Models and inputs}
\centering
\begin{tabular}[t]{ l r l r }
  \toprule
  Name used & Input image size & Base model name & Params \\ [0.25ex]
  \midrule
  tiny-640     & 640x640 pixels   & YOLOv7-tiny & 6.2M   \\
  full-640     & 640x640 pixels   & YOLOv7      & 36.9M  \\
  tiny-1280    & 1280x1280 pixels & YOLOv7-tiny & 6.2M   \\
  full-1280    & 1280x1280 pixels & YOLOv7      & 36.9M  \\
  \bottomrule
\end{tabular}
\label{table:models}
\end{table}

The hardware used for training was an NVIDIA A100 40GB GPU in the VUW Rāpoi HPC cluster \cite{vuw_raapoi_docs}. The hardware used for inference was an NVIDIA GTX 1660 Ti laptop GPU. 

\section{Buoy Detection Experiments}

\subsection{Evaluation Metrics}

The main metrics used in this study are the inference frames per second (FPS), to evaluate processing speed, along with the mean average precision (mAP) to evaluate detection accuracy.

\subsubsection{Inference time} 

The time the inference process takes when detecting the buoys on a given image.

\subsubsection{Inference FPS}

The number of images a model can detect buoys for per second. Ideally, this should be higher than the video FPS for processing videos in real-time.

\subsubsection{IoU}

Intersection over union (IoU) is the proportional area of overlap of two bounding boxes, normally the ground truth and prediction bounding boxes \cite{hui_map_2018}.

\subsubsection{mAP}

Mean average precision (mAP) takes the area under the precision versus recall curve, from 0.5 IoU to 0.95 IoU, in increments of 0.05 steps, which is averaged over all IoUs, and all classes (referred to as mAP@[.5:.05:.95] or shortened to mAP) \cite{hui_map_2018}. The mean average precision at 0.5 IoU (mAP@0.5) is more lenient, focusing on mAP at 0.5 IoU.

\subsection{Performance of Fine-tuned Models}

\begin{figure}[htbp]
  \centerline{\includegraphics[width=9cm]{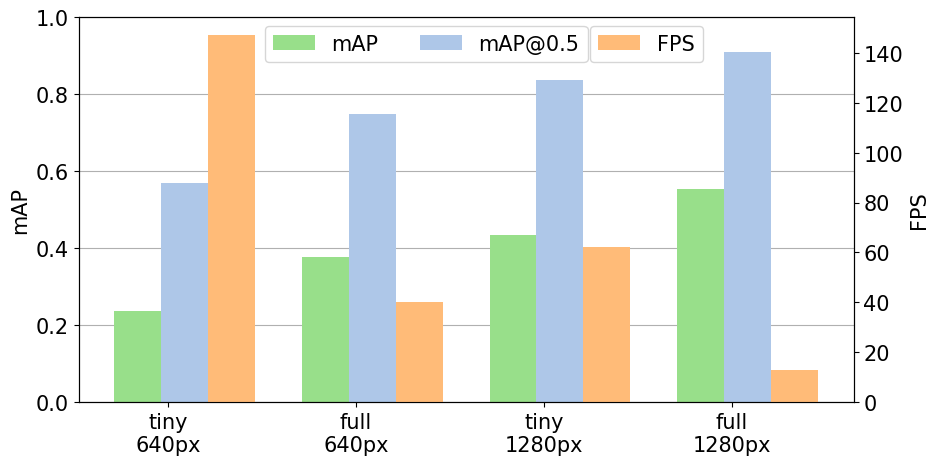}}
\caption{Each model variation shows mAP, mAP@0.5, and inference FPS. Higher is better for all displayed metrics.}
\label{fig:all_metrics}
\end{figure}

Results for inference time and FPS are presented in Table \ref{table:time_metrics}, and FPS is visualised in Fig. \ref{fig:all_metrics}. As observed, the inference FPS of the tiny-640 model was much faster than the others at about 147 FPS, and the full-1280 model was much slower at about 13 FPS. The cameras used for video on the mussel farm shoot at 15 FPS, so all models except the full-1280 model could process images at a feasible real-time speed.

\begin{table}[htbp]
	\caption{Time metrics}
	\centering
	\begin{tabular}[t]{ l c r r r }
		\toprule
		Model & \thead{Inference\\(ms)} & \thead{FPS} & \thead{Memory\\(MB)} \\ [0.25ex]
		\midrule
		tiny-640  & 6.8       & 147.1  & 972 \\
		full-640  & 24.5      & 40.8   & 1324 \\
		tiny-1280 & 16.1      & 62.1   & 1108 \\
		full-1280 & 75.4      & 13.3   & 1672 \\
		\bottomrule
	\end{tabular}
	\label{table:time_metrics}
\end{table}

As shown in Table \ref{table:time_metrics}, the maximum memory used for each model during inference varied from about 1GB to about 1.7GB. This range of memory should be acceptable for most modern GPUs that would be used in real world applications.

Table \ref{table:performance_metrics} presents performance metrics for each model evaluated on the all combined test set. Test parameters were non-maximum suppression (NMS) IoU threshold of 0.65 and confidence threshold of 0.001. In terms of mAP and mAP@0.5, the full-1280 model had the highest performance, and the tiny-640 model had the lowest performance, as would be expected. This is further illustrated in Fig. \ref{fig:all_metrics}. An interesting result is that the tiny-1280 model outperforms the full-640 model, which shows that increasing the input image size from 640 pixels to 1280 pixels has a bigger positive impact on performance than increasing the model size from YOLOv7-tiny to YOLOv7. The likely explanation for this is the larger image size increases the ability of both YOLOv7-tiny and YOLOv7 to detect smaller buoys in the images, which could be lost in smaller images.

\begin{table}[htbp]
\caption{Performance metrics}
\centering
\begin{tabular}[t]{ l c c c c c c }
  \toprule
  Model & F1 & Precision & Recall & mAP & \thead{mAP\\@0.5} \\ [0.25ex]
  \midrule
  tiny-640  & 0.579 & 0.684 & 0.502 & 0.237 & 0.569 \\
  full-640  & 0.726 & 0.823 & 0.649 & 0.378 & 0.748 \\
  tiny-1280 & 0.785 & 0.816 & 0.756 & 0.435 & 0.836 \\
  full-1280 & 0.865 & 0.890 & 0.841 & 0.554 & 0.908 \\
  \bottomrule
\end{tabular}
\label{table:performance_metrics}
\end{table}

Fig. \ref{fig:all_metrics} summarising mAP, mAP@0.5, and inference FPS, shows that of the four models, YOLOv7-tiny with 1280px input images achieves the best balance of these metrics. Its mAP and mAP@0.5 are slightly lower than YOLOv7 with 1280px input images but the inference FPS is much higher.

\begin{figure*}[htbp]
  \centerline{\includegraphics[width=18cm]{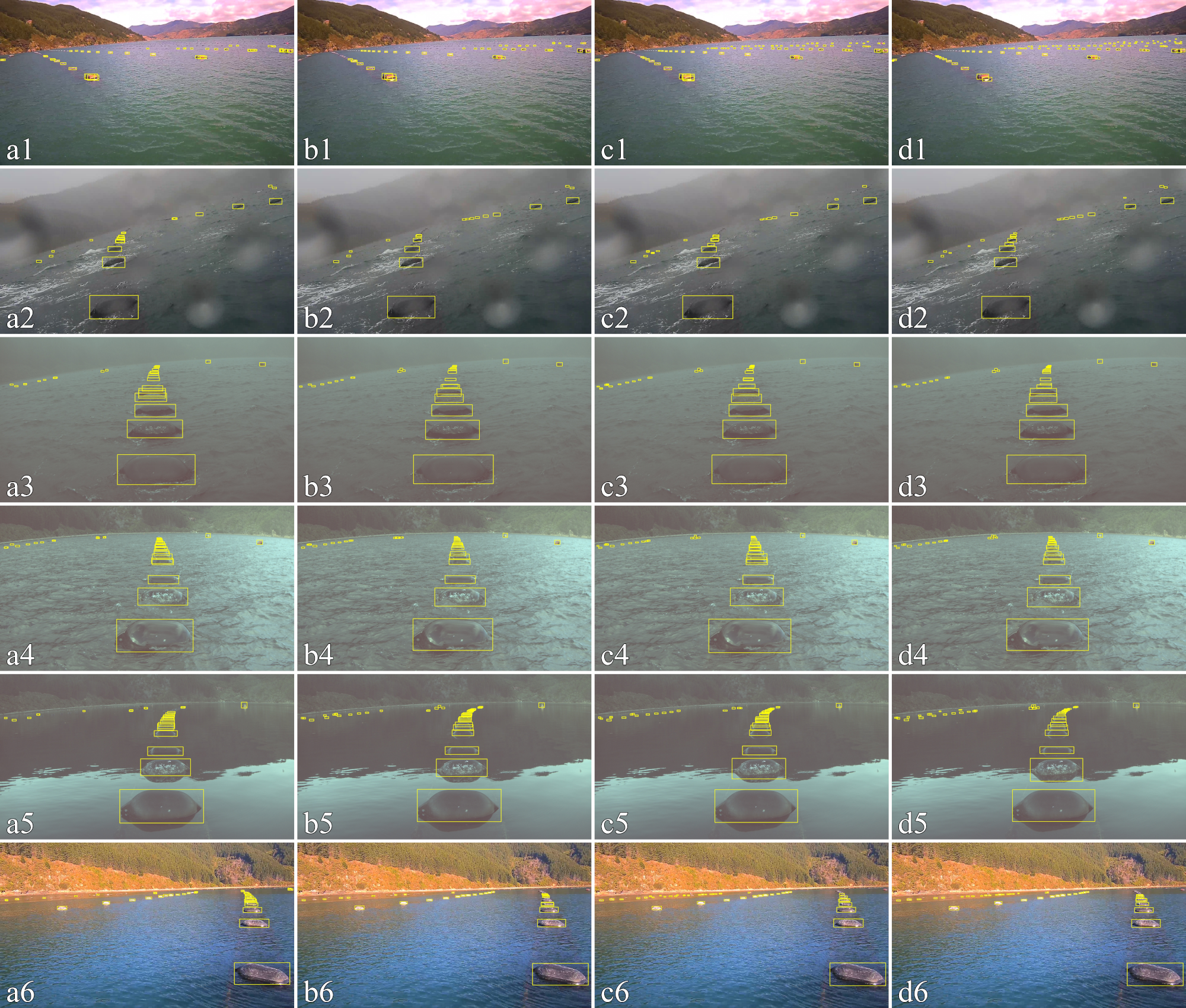}}
  \caption{A different model is shown in each column: a. tiny-640, b. full-640, c. tiny-1280, d. full-1280. Various example images demonstrating a variety of scenarios are shown in each row as: 1. Many small buoys, 2. Choppy water, 3. Foggy conditions, 4. Rough conditions, 5. Calm conditions with distinct shadows on the water, 6. Calm conditions }
  \label{fig:sample_images}
\end{figure*}

Detection sample images of each model across a range of conditions are shown in Fig. \ref{fig:sample_images}. All models predicted the large buoys well. The models with the larger 1280px images as input detected more of the smaller buoys in the background, which is most apparent in row 1 from Fig. \ref{fig:sample_images}. An interesting anomaly in row 6 from Fig. \ref{fig:sample_images} is while the tiny-640 model misses many small buoys at the images left, at the right it detects many buoys in the distance that none of the other models detected. However, based on the much lower mAP results in Fig. \ref{fig:all_metrics} for the tiny-640 model, it is likely a rare occurrence.

In practicality, all models perform well if the concern is only large buoys in the foreground, however, if small buoys in the background are needed then the 1280px models look most acceptable, with the exception that the full-1280 model may run too slow for real-time detection.

\subsection{Ablation study}

The models from Table \ref{table:models} were trained and tested on the various datasets in Table \ref{table:datasets} to see YOLOv7's performance on the data variations. The adverse weather data was also tested on all models to see how well they handled adverse weather conditions. Performance metrics for models trained and tested on the boat camera dataset, buoy camera (high res) dataset, and buoy camera (low res) dataset are shown in Table \ref{table:performance_metrics_datasets}.

\begin{table}[htbp]
\caption{Results for each dataset}
\centering
\begin{tabular}[t]{ l c c c c c }
  \toprule
  Model & F1 & Precision & Recall & mAP & \thead{mAP\\@0.5} \\ [0.25ex]
  \midrule
  \multicolumn{6}{l}{\textbf{Results of Boat camera trained models}} \\
  tiny-640  & 0.675 & 0.791 & 0.588 & 0.332 & 0.667 \\
  full-640  & 0.667 & 0.769 & 0.589 & 0.324 & 0.655 \\
  tiny-1280 & 0.844 & 0.875 & 0.816 & 0.546 & 0.900 \\
  full-1280 & 0.843 & 0.877 & 0.812 & 0.542 & 0.898 \\
  \multicolumn{6}{l}{\textbf{Results of Buoy camera (high res) trained models}} \\
  tiny-640  & 0.585 & 0.560 & 0.613 & 0.244 & 0.594 \\
  full-640  & 0.574 & 0.607 & 0.545 & 0.234 & 0.574 \\
  tiny-1280 & 0.793 & 0.801 & 0.785 & 0.413 & 0.841 \\
  full-1280 & 0.804 & 0.829 & 0.780 & 0.425 & 0.850 \\
  \multicolumn{6}{l}{\textbf{Results of Buoy camera (low res) trained models}} \\
  tiny-640  & 0.515 & 0.574 & 0.467 & 0.206 & 0.495 \\
  full-640  & 0.522 & 0.569 & 0.483 & 0.206 & 0.490 \\
  tiny-1280 & 0.789 & 0.798 & 0.780 & 0.456 & 0.847 \\
  full-1280 & 0.795 & 0.839 & 0.755 & 0.461 & 0.848 \\
  \bottomrule
\end{tabular}
\label{table:performance_metrics_datasets}
\end{table}

In terms of mAP, the model trained on the boat camera images performed best. A possible reason is the boat camera dataset has much more training data than the buoy camera datasets. The performance of models trained with all combined data was better than those trained with boat camera data on the YOLOv7 models, but worse on the YOLOv7-tiny models.

Results for the adverse weather test data are shown in Table \ref{table:performance_metrics_adverse_weather} (HR is high res, LR is low res). Results are only shown for the YOLOv7 with 1280 pixel size input (full-1280) model for brevity as results for the other models were relatively similar.

\begin{table}[htbp]
\caption{Results of Adverse weather}
\centering
\begin{tabular}[t]{ l c c c c c c }
  \toprule
  \thead{Dataset model\\was trained on} & F1 & Precision & Recall & mAP & \thead{mAP\\@0.5} \\ [0.25ex]
  \midrule
  Boat cam      & 0.545 & 0.685 & 0.453 & 0.298 & 0.501 \\
  Buoy cam (HR) & 0.474 & 0.543 & 0.421 & 0.189 & 0.441 \\
  Buoy cam (LR) & 0.772 & 0.803 & 0.744 & 0.477 & 0.834 \\
  All combined  & 0.826 & 0.836 & 0.816 & 0.571 & 0.886 \\
  \bottomrule
\end{tabular}
\label{table:performance_metrics_adverse_weather}
\end{table}

In terms of mAP, performance was lower for the adverse weather test set against the boat camera and high-resolution buoy camera models. However, performance was relatively better against the low-resolution buoy camera models and all combined dataset models compared with other models. This was expected because the adverse weather test images were sampled from the same data source as the low-resolution buoy camera dataset, which is also included in the combined dataset.

\begin{figure}[htbp]
  \centerline{\includegraphics[width=9cm]{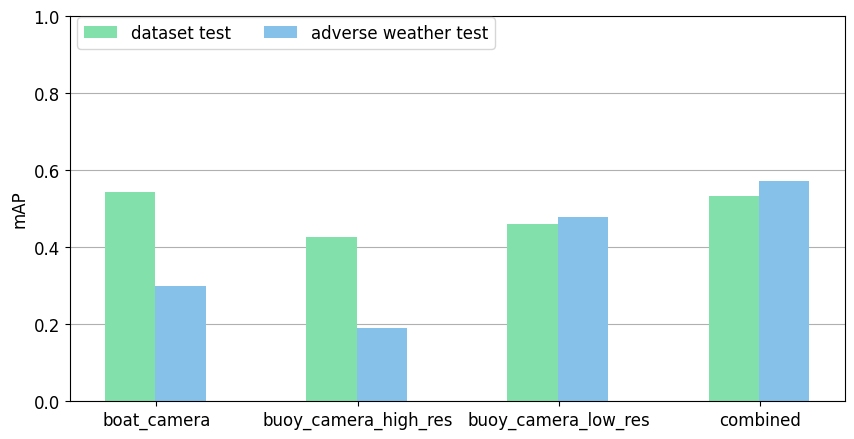}}
  \caption{YOLOv7 with 1280px input model trained and tested on various datasets and also tested on the adverse weather dataset shows mAP@[0.5:0.05:0.95].}
  \label{fig:map_adverse_compared}
\end{figure}

Fig. \ref{fig:map_adverse_compared} shows the mAP of the full-1280 models trained using each dataset, evaluated with their own test set and the adverse weather test set. For the results, we can see that the low-resolution buoy camera model and the combined model outperform the other models by a large margin when evaluated on the adverse weather test dataset.

It's noteworthy that we did not need additional preprocessing of the adverse weather images before feeding into the YOLO models as other studies \cite{liu_image-adaptive_2022,wang_denet_2023} have used before. This is because YOLOv7 has improved over YOLOv3 used in the previous studies \cite{liu_image-adaptive_2022,wang_denet_2023} for detecting objects in adverse weather conditions as reported by Haimer et al. \cite{haimer_yolo_2023}.

\section{Limitations}

The empirical parameters in Table \ref{table:training_params} used for training and evaluation may not be optimal for the specific scenario of buoy detection in an ocean environment. Given the sizeable parameter set, it is an ample-sized task to find optimal values, which can be addressed in future work.

The GPU used for inference testing was an approximation to a target portable GPU by using an NVIDIA GTX 1660 Ti laptop GPU. In a real-world scenario, a dedicated portable GPU will be more applicable for our application, such as the robotics device NVIDIA Jetson Nano. 

\section{Conclusions and Future Work}

This study developed a deep learning model for buoy detection to automate mussel farm supervision. A transfer learning-based pipeline was proposed and the effectiveness of the resultant model was exhaustively evaluated, including robustness under adverse weather conditions. The model displayed commendable accuracy and efficient inference speed when evaluated with real world resource constraints, suggesting its effectiveness for real time buoy detection in mussel farm video feeds.

Further work would be to accommodate the outcome of this study into real usage. These developed deep learning models could be deployed as part of an automated mussel farm monitoring pipeline for detecting drifting or sinking buoys to assist its maintenance operations. Additionally, further investigation could be undertaken to find the optimal hyperparameters and image augmentations and improve small object detection in sea images.

\section*{Acknowledgments}

Thank you Dylon Zeng, Ying Bi, and Ivy Liu for joining the discussions and providing suggestions during this research.

Our research received financial support from the Science for Technological Innovation Challenge (SfTI) fund, under contract number 2019-S7-CRS, and the MBIE Data Science SSIF Fund, under contract RTVU1914.

\bibliographystyle{IEEEbib}
\bibliography{refs}

\end{document}